\documentclass{article} % For LaTeX2e
\usepackage{amsthm}
\usepackage{amsthm}
\usepackage{iclr2026_conference,times}
\usepackage{graphicx}
\usepackage{amsmath,amsfonts,bm}

\def\eqref#1{equation~\ref{#1}}
\def\1{\bm{1}}

\DeclareMathAlphabet{\mathsfit}{\encodingdefault}{\sfdefault}{m}{sl}
\SetMathAlphabet{\mathsfit}{bold}{\encodingdefault}{\sfdefault}{bx}{n}

\DeclareMathOperator*{\argmax}{arg\,max}

\usepackage{hyperref}
\usepackage{url}
\usepackage{algorithm}
\usepackage{algorithmic}
\usepackage{booktabs}
\usepackage{amsthm}
\newtheorem{definition}{Definition}
\newtheorem{remark}{Remark}
\newtheorem{proposition}{Proposition}

\title{Action Shapley: A Training Data Selection Metric for World Model in Reinforcement Learning
}

\iclrfinalcopy % Uncomment for camera-ready version, but NOT for submission.
\usepackage{amsthm}
\newtheorem{corollary}{Corollary}

\author{
Rajat Ghosh \\
Nutanix Inc. \\
\texttt{rajat.ghosh@nutanix.com}
\And
Debojyoti Dutta \\
Nutanix Inc. \\
\texttt{debojyoti.dutta@nutanix.com}
}

\begin{document}

\maketitle

\begin{abstract}
Numerous offline and model-based reinforcement learning systems incorporate world models to emulate the inherent environments. A world model is particularly important in scenarios where direct interactions with the real environment is costly, dangerous, or impractical. The efficacy and interpretability of such world models are notably contingent upon the quality of the underlying training data. In this context, we introduce Action Shapley as an agnostic metric for the judicious and unbiased selection of training data. To facilitate the computation of Action Shapley, we present a randomized dynamic algorithm specifically designed to mitigate the exponential complexity inherent in traditional Shapley value computations. Through empirical validation across five data-constrained real-world case studies, the algorithm demonstrates a computational efficiency improvement exceeding 80\% in comparison to conventional exponential time computations. Furthermore, our Action Shapley-based training data selection policy consistently outperforms ad-hoc training data selection.
\end{abstract}

\section{Introduction}
\label{introduction}

\begin{figure}[ht]
	\vskip 0.2in
	\begin{center}
    \includegraphics[width=\columnwidth]{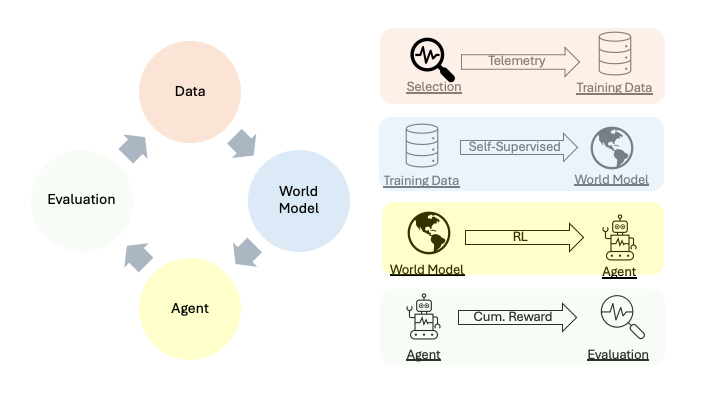}
		\caption{The RL training loop. (a) the training data is selected for the world model development. (b) the collected training data is used to train (update) the world model by a self-supervised algorithm.  (c) With the world model, the agent updates its policy by using an RL algorithm. (d) The agent policy is evaluated. If needed, the agent goes back to step (a) for further improvement. }
		\label{fig:schematic} 
	\end{center}
	\vskip -0.2in
\end{figure}

We introduce Action Shapley, a training data selection metric for a world model \citep{kaiser2019model_world_model, ha2018world_model, doll2012ubiquity_world_model} to be consumed by a model-based reinforcement learning (RL) agent  \citep{sutton2018reinforcement}. It is inspired by the Shapley value \citep{shapley_value,shapley_value_computation}  introduced by L.S. Shapley.  Although prior works have proposed Shapley based data evaluation in supervised learning  \citep{data_shapley} and interpretability \cite{SHAP}, no prior work has demonstrated the application of Shapley for reinforcement learning. 
Compared to supervised learning, reinforcement learning brings additional complexity due to its dynamic learning and stability problems such as the deadly triad  \citep{sutton2018reinforcement}. Figure \ref{fig:schematic} shows the end-to-end RL training workflow including training data selection, world model training, agent training based on a RL algorithm, and agent evaluation by measuring the cumulative reward score. Action Shapley measures the marginal contribution of a training data point for this end-to-end workflow.

In this paper, we examine the scenarios where an RL agent is tasked with dynamic control system management. 
Model-based reinforcement learning (MBRL) offers a powerful approach to address dynamic control challenges by combining the sequential decision-making of reinforcement learning (RL) with the world models for system dynamics. By building accurate world models of the underlying system dynamics, MBRL can predict how the system will respond to different control inputs, enabling it to compute an optimal policy  that maximize a cumulative reward score. This paradigm finds applications in cloud computing  \citep{vm_sizing, load_balancing, db_config, autopilot, rl_vmsizing, rl_loadbalancing, rl_dbtuning, rl_k8s}, commercial cooling \citep{deepmind_cooling}, robotics \citep{mbrl_robotics}, autonomous vehicles \citep{controls_car_DRL}, and industrial automation \citep{RL_PID_industry}, where precise control is essential. MBRL not only enhances the adaptability of control strategies in complex, dynamic environments but also provides a mechanism to incorporate safety constraints \citep{rl_safety} and handle system uncertainties. Furthermore, MBRL can significantly reduce the amount of real-world data required for training \citep{data_efficiency_MBRL}, making it an attractive choice for applications where experimentation can be costly or impractical.  Despite its benefits, the adoption of RL in dynamic control systems is stymied by data related practical issues such as data sparsity \citep{kamthe2018data}, noisy environment \citep{dulac2019challenges}, partial observability \citep{futoma2020popcorn}, and irregular data sampling \citep{irr_sampling}.  The aforementioned issues in data quality therefore validate the usefulness of Action Shapley.

 \section{Action Shapley}
 \label{methodology}

 We consider a stochastic environment with states $s \in S \subseteq \mathbb{R}^{p} $, actions $a \in A \subseteq \mathbb{R}^{q} $, and additive noise $\omega_{n} \in \mathbb{R}^{p}$. The resulting transition dynamics is represented as:
 
\begin{equation}
	\label{eq:world_model}
	s_{n+1} = f(s_n, a_n) + \omega_n
\end{equation}

where, $f: S \times A \rightarrow S $. For  tractability, we assume $f$ to be Lipschitz continuous. We aim to approximate $f$ with a world model (often represented with a neural network). Self-supervised algorithms are generally used for a world model computation \citep{ss_world_model}. It aims to understand and simulate the environment by training on unlabeled data, learning patterns and representations from it.

The goal of the RL agent is control the stochastic system (in Eq. \ref{eq:world_model}) optimally in an episodic setting over a finite time horizon $N$. To control the system, we use a deterministic policy $\pi_{n}: S \rightarrow A$ from  a set of $\Pi$ that selects actions $a_n = \pi_{n}(s_n)$ given the current state. For ease of notation, we assume that the system is reset to a known state $s_{0}$ at the end of each episode, that there is a known reward function, $r: S \times A \rightarrow \mathbb{R}$. For any dynamic world model $f: S \times A \rightarrow S$, the performance of a policy $\pi$ is the total reward collected during an episode in expectation over the transition noise, $\omega$:

\begin{equation}
	\label{eq:objective}
\begin{aligned}
	J(f, \pi) = \mathbb{E}_{\omega_{0: N-1}} \left[\sum^{N}_{n=0}r(s_n, \pi(s_n))  \vert s_{0} \right], \\ 
	\ni s_{n+1} = f(s_n, a_n) + \omega_n
	\end{aligned}
\end{equation}

Thus, we aim to find the optimal policy $\pi^{*}$ for $f$ with the following equation:

\begin{equation}
	\label{eq:optimal_policy}
	\pi^{*} = \argmax_{\pi \in \Pi} J(f, \pi)
\end{equation}

The concept of Action Shapley emerges as a metric designed to gauge the differential contribution of each data point within a dataset $\mathcal{D}$ to the resultant RL agent. It is relevant in the context of training data selection and evaluation (steps (a) and (d) in RL training loop as shown in Figure \ref{fig:schematic}). This differential contribution is quantified through a valuation function, denoted as $\mathcal{U}$. For this paper, $\mathcal{U}$ is set as the cumulative reward garnered by the RL agent.

\begin{definition}
\label{def:AS}

In a general formulation, Action Shapley is expressed as $\phi(\mathcal{D}; \mathcal{U} )$. To ensure equitable and robust valuation, we posit that $\phi$ adheres to the properties of nullity, symmetry, and linearity. Under these assumptions, the Action Shapley value pertaining to a specific training data point $\{k\}$ is delineated by Equation \ref{eq:shapley}, as per the established literature \citep{data_shapley, shapley_value_computation, shapley_sampling, shapley_value}.

\begin{equation}
	\label{eq:shapley}
	\phi_{k}=C_{f}\sum_{d \subseteq \mathcal{D} \backslash \{k\} } \frac{\mathcal{U}(d \cup \{k\}) - \mathcal{U}(d)}{\binom{n-1}{\vert{d}\vert}}
\end{equation}

where, $C_{f}$ is an empirical constant. While Equation \ref{eq:shapley}'s numerator gauges the distinct contribution of training data point ${k}$ with the dataset, $d$, it's denominator acts as the weight for the dataset, $d$. By summing over all possible dataset combinations, it provides a robust measure of the total differential contribution of ${k}$, surpassing the traditional Leave-One-Out (LOO) approach in marginal contribution determination and deduplication.

\end{definition}

\begin{remark}
	  In cases where the RL agent fails to achieve its goal for a training dataset, $d$, we assign $\mathcal{U}(d) \to \text{null}$. 
\end{remark}

\begin{definition}
\label{def:cutoff}
A successful RL agent's world model, $f: S \times A \rightarrow S $, necessitates a threshold number of training data points. We defined it as the cut-off cardinality ($\theta$). 
\end{definition}

\begin{proposition}
We leverage the cut-off cardinality to circumvent the default exponential-time complexity and introduce a randomized dynamic algorithm denoted as Algorithm \ref{alg:ASC} in Appendix.
\end{proposition}

 \begin{remark}
 Algorithm \ref{alg:ASC} employs a top-down accumulator pattern. Initialization involves considering the set encompassing all training data points, followed by a breadth-first traversal of the power set tree of the training data. As depicted in Algorithm \ref{alg:ASC}, the structure involves a nested $While$ loop syntax. The outer $While$ loop iterates over training data cardinalities in a descending order, while the inner $While$ loop iterates over all possible training data subsets for the specified cardinality. Within the inner loop, a check for RL agent failure is conducted (line 8 in Algorithm \ref{alg:ASC}). In the event of a failure, the failure memoization accumulator is incremented by one (line 10 in Algorithm \ref{alg:ASC}). The termination condition is met when the failure memoization accumulator, denoted as $mem$, reaches a user-defined threshold parameter, $\epsilon$. Upon termination, the cut-off cardinality (line 12) is set to be the cardinality where the failure condition (line 8) is first satisfied, incremented by one. If the failure condition is not met, the marginal contribution of the training data ($k$) concerning the subset $d$ is added to the accumulation variable $sum$. This process of accumulating differential contributions continues until the termination condition is satisfied (line 17 in Algorithm \ref{alg:ASC}).
 \end{remark}

\begin{definition}
	\label{def:global_cutoff}
	Each training action is associated not only with its respective Action Shapley value but also possesses a unique \textit{cut-off cardinality} value. The global \textit{cut-off cardinality}, defined as the maximum among the \textit{cut-off cardinality} values corresponding to distinct training data points, serves as an upper computational limit for Algorithm \ref{alg:ASC}. 
\end{definition}

The computational complexity of Algorithm \ref{alg:ASC} is situated between $\mathcal{O}(2^n)$ ($n$ is the total number of training data points), denoting the worst-case scenario, and $\mathcal{O}(\epsilon)$, reflecting the best-case scenario. The worst-case arises when the termination condition (line 11) is not met, leading to the exhaustive traversal of the entire combination tree. Conversely, the best-case occurs when the first $\epsilon$ evaluations fail, promptly reaching the termination condition (line 11).

\begin{corollary}
The best-case scenario implies the indispensability of the training data point ${k}$, indicating that all agents fail to achieve the goal without the inclusion of data point ${k}$ in training. In such instances, data point ${k}$ is termed indispensable. It is noteworthy that a comparable approach involving the \textit{cut-off cardinality} has been employed successfully in the literature \citep{shapley_sampling}.
\end{corollary}

\begin{proposition}
	Combining these two extremes of $\mathcal{O}(2^n)$ and $\mathcal{O}(\epsilon)$, the performance of Algorithm \ref{alg:ASC}  can be represented as a ratio of the exponential of global cut-off cardinality and the exponential of the total number of training data points subtracted from $1$, as shown in Equation \ref{eq:comp_effort}. 
	\begin{equation}
		\label{eq:comp_effort}
		P_{comp} = 1 - \frac{2^{\theta_{k_{max}}}}{2^n}
	\end{equation}
\end{proposition}

\begin{proposition}
\label{prop:proposal}
	 A set exhibiting a higher average Action Shapley value outperforms a set with a lower average value in producing a high-performance RL agent. We propose an ideal training dataset should comprise no more training data points than the specified global cut-off cardinality and should have the highest possible average Action Shapley value. Low sample complexity improves computational efficiency. However, a minimum number of training data points is essential for a successful reinforcement learning (RL) agent. The proposed policy is underpinned by a theoretically sound foundation as an explanatory model for Shapley-based Additive Feature Attribution Methods \cite{SHAP}. 
\end{proposition}

\section{Data Collection and Implementation}
\label{data collection and implementation}

We have implemented five distinct case studies, as follows, to substantiate the effectiveness of Action Shapley.

\begin{itemize}
	\item \textit{VM Right-Sizing}: RL agent to  adjust vCPU count and memory size for a VM  in order to bring its p50 CPU usage below a pre-assigned threshold $(90\%)$.
	\item  \textit{Load Balancing}: RL agent to adjust CPU and memory worker counts for a VM in order to bring its p5 CPU usage below a pre-assigned threshold $(70\%)$. 
	\item  \textit{Database Tuning}: RL agent to adjust vCPU count and memory size for a database in order to bring its p90 CPU usage below a pre-assigned threshold $(25\%)$.
	\item  \textit{Kubernetes (k8s) Management}: RL agent to maximize write rates on a distributed database running on a Kubernetes cluster while keeping the p99.9 latency below a pre-assigned threshold $(100 ms)$. 
	\item  \textit{Data Center Cooling Management}: RL agent to adjust cooler temperature set point and pump pressure set point to keep p99.9 CPU temperature just below a  pre-assigned threshold $(65 ^{0}C)$.
\end{itemize}

 In the investigations concerning VM right-sizing and load balancing, AWS EC2 instances have been deployed, incorporating the \emph{Stress} tool to facilitate dynamic workload simulations. In the context of the database tuning case study, we have employed a Nutanix AHV-based virtualized infrastructure, utilizing the \emph{HammerDB} tool for dynamic SQL load simulation. The case study focusing on Kubernetes management entails the utilization of a Nutanix Kubernetes (K8s) engine cluster, with \emph{cassandra-stress} employed for workload simulations. The case study focusing on data center cooling management uses a test data center environment with a standardized C++ code for workload simulation. In each case study, the training dataset is comprised of a collection of time series corresponding to different configurations. These time series are measured over a 24-hour period at a 1-minute sampling rate. It is important to note that, across all case studies, the error bound ($\epsilon$) is consistently fixed at 1.

Action Shapley is agnostic to the specific instantiations of the RL algorithm and the world model algorithm for the environment. For the world model, we use a radial basis function (RBF) network \citep{he2019landslide}, along with autoencoder-based pre-training \citep{nature_2019_unsupervised}, for the $(state_{prev}, action_{prev}) \rightarrow (state_{cur}, reward_{cur})$ mapping.  Without loss of generality, we choose two different types of RL algorithms: SAC-PID \citep{sac_pid} and PPO-PID \citep{ppo_pid}. The PID loop, as shown in Equation \ref{eq:pid}, is used as an RL action update policy based on the error term ($= \left(\text{threshold} - \text{aggregated state statistic}\right)$), the time step ($(\delta t)$), and learning parameters, $[k_{p}, k_{i}, k{d}]$.

\begin{gather}
	\label{eq:pid}
	a_{i} = a_{i-1} + 
	\begin{bmatrix}
		kp & ki & kd
	\end{bmatrix}
	\cdot
	\begin{bmatrix}
		e & (\delta t) e & \frac{e}{\delta t} \\
	\end{bmatrix}
\end{gather}

For the VM right-sizing case study, we have five different data points: $\langle a_1:(2,2), a_2:(2,4), a_3:(2,8), a_4:(4,16), a_5:(8,32) \rangle$. Each pair represents $(\text{vCPU count}, \text{Memory Size in GB})$. The training dataset consists of five time series of CPU usage, each with 1440 data points.  While p50 is used for the aggregated state statistic, the goal of the RL agent is to bring the p50 CPU usage below $90\%$. The starting action for the RL loop is $(6, 14)$.  The error bound, $\epsilon$, is set to 1.

For the load balancing case study, we have five different data points:   
$\langle w_1:(8,16), w_2:(8,12), w_3:(8,2), w_4:(1,2), w_5:(1,16) \rangle$. Each pair represents $(\text{\# of CPU workers}, \text{\# of memory workers}).$ The training dataset consists of five time series of CPU usage, each with 1440 data points. While p5 is used for the aggregated state statistic, the goal of the RL agent is to bring the p5 CPU usage below $70\%$. The starting action for the RL loop is $(5, 10)$.   The error bound, $\epsilon$, is set to 1.

For the database tuning case study, we have six different data points: $\langle p_1: (1,1), p_2:(4,4), p_3:(6,3), p_4:(8,4), p_5:(8,8), p_6:(10,10) \rangle$.  Each pair represents $(\text{vCPU count}, \text{Memory Size in GB})$. The training dataset consists of six time series of CPU usage, each with 1440 data points. While p90 is used for the aggregated state statistic, the goal of the RL agent is to bring the p90 CPU usage below $25\%$. The starting action for the RL loop is $(5, 2)$.  The error bound, $\epsilon$, is set to 1.

In the Kubernetes management case study, we employ 15 different data points $\langle r_{1}: (1 \times 10^6, 10), r_{2}: (1 \times 10^6, 25),  r_{3}: (1 \times 10^6, 50),  r_{4}: (1 \times 10^6, 75),  r_{5}: (1 \times 10^6, 100), r_{6}: (2 \times 10^6, 10), r_{7}: (2 \times 10^6, 25),  r_{8}: (2 \times 10^6, 50),  r_{9}: (2 \times 10^6, 75),  r_{10}: (2 \times 10^6, 100), r_{11}: (3 \times 10^6, 10), r_{12}: (3 \times 10^6, 25),  r_{13}: (3 \times 10^6, 50),  r_{14}: (3 \times 10^6, 75),  r_{15}: (3 \times 10^6, 100) \rangle$.
Each training configuration is denoted by a pair representing the tuple ($\text{write rate}$, $\text{thread count}$). The training dataset encompasses 15 time series data sets, each comprising 1440 data points, measuring response latency. The aggregated state statistic is determined by the p99.9 metric, and the primary objective of the RL agent is to reduce the p99.9 latency to below 100ms. The RL loop commences with an initial action of $(2.9 \times 10^6, 95)$.  The error bound, $\epsilon$, is set to 1.

In the data center cooling management case study, we employ 12 different data points $\langle c_{1}: (29, 4), c_{2}: (29, 7),  c_{3}: (29,10),  c_{4}: (25, 4),  c_{5}: (25, 7), c_{6}: (25, 10), c_{7}: (21, 4),  c_{8}: (21, 7),  c_{9}: (21, 10),  c_{10}: (17, 4), c_{11}: (17, 7), c_{12}: (17, 10) \rangle$ (units are in $^{0}C$, psi).
Each training configuration is denoted by a pair representing the tuple ($\text{cooler temperature set point}$, $\text{pump pressure set point}$). The training dataset encompasses 12 time series data sets, each comprising 1440 data points, measuring CPU temperatures against every training data points. The aggregated state statistic is determined by the p99.9 metric, and the primary objective of the RL agent is to increase the p99.9 CPU temperature to just below 65 $^{0}C$ at the most cost-efficient cooling setpoint. The RL loop commences with an initial cooling setpoint of $(29, 4)$ .  The error bound, $\epsilon$, is set to 1.

\section{Results}

\begin{table}[t]
	\caption{Action Shapley (AS) for VM right-sizing case study for two different RL algorithms, SAC-PID and PPO-PID}
	\label{tab:as_c1}
	\vskip 0.15in
	\begin{center}
		\begin{small}
			\begin{sc}
				\begin{tabular}{lcccr}
					\toprule
					Data set & AS SAC-PID & AS PPO-PID \\
					\midrule
					$a_1 $ & -3.96 & -3.90   \\
					$\bf{a_2}$ & \bf{-1.84} & \bf{-1.81}    \\
					$a_3$ & ind. & ind. \\ 
					$a_4$ & ind. & ind.\\
					$a_5 $ & ind. & ind.\\
					\bottomrule
				\end{tabular}
			\end{sc}
		\end{small}
	\end{center}
	\vskip -0.1in
\end{table}

\begin{table}[t]
	\caption{Action Shapley (AS) for load balancing case study for two different RL algorithms, SAC-PID and PPO-PID}
	\label{tab:as_c2}
	\vskip 0.15in
	\begin{center}
		\begin{small}
			\begin{sc}
				\begin{tabular}{lcccr}
					\toprule
					Data set & AS SAC-PID & AS PPO-PID \\
					\midrule
					$w_1$ & 0.57 & 0.57    \\
					$w_2$ &  1.02  & 1.10  \\
					$\bf{w_3}$ & \bf{8.31}  & \bf{8.27} \\
					$w_4$ & 3.61 & 3.61\\
					$w_5$ & 5.55 & 5.51\\
					\bottomrule
				\end{tabular}
			\end{sc}
		\end{small}
	\end{center}
	\vskip -0.1in
\end{table}

\begin{table}[t]
	\caption{Action Shapley (AS) for database tuning case study for two different RL algorithms, SAC-PID and PPO-PID}
	\label{tab:as_c3}
	\vskip 0.15in
	\begin{center}
		\begin{small}
			\begin{sc}
				\begin{tabular}{lcccr}
					\toprule
					Data set & AS SAC-PID & AS PPO-PID \\
					\midrule
					$p_1$ &  Ind. & ind.    \\
					$p_2$ &  1.11 & 1.12   \\
					$\bf{p_3}$ &  \bf{3.42}   & \bf{3.37}\\
					$p_4$ &  1.75 & 1.72\\
					$p_5$ &  0.19 & 0.21\\
					$p_6$ &  -0.23 & -0.22\\
					\bottomrule
				\end{tabular}
			\end{sc}
		\end{small}
	\end{center}
	\vskip -0.1in
\end{table}

\begin{table}[t]
	\caption{Action Shapley (AS) for Kubernetes management case study for two different RL algorithms, SAC-PID and PPO-PID}
	\label{tab:as_c4}
	\vskip 0.15in
	\begin{center}
		\begin{small}
			\begin{sc}
				\begin{tabular}{lcccr}
					\toprule
					Data set & AS SAC-PID & AS PPO-PID \\
					\midrule
					$r_1$   & -0.7 & 0.68      \\
					$r_2 $   & 0.53 & 0.54      \\
					$r_3 $   & 0.61 & 0.62   \\
					$r_4$   & -0.13 & -0.12     \\
					$r_5$   & 0.12 & 0.11      \\
					
					$r_6$   & -0.7 & -0.7  \\
					$r_7$   & -0.24 & -0.25 \\
					$r_8$   & 0.65 & 0.65 \\
					$r_9$   & 0.42 & 0.42\\
					$r_{10}$   & 0.08 & 0.07 \\
					
					$r_{11}$   & -1.16 & -1.17 \\
					$r_{12}$   & -0.25 & -0.24 \\
					$\bf{r_{13}}$   & \bf{0.77} & \bf{0.77}  \\
					$r_{14}$   & -0.31 & -0.31 \\
					$r_{15}$    & 0.019 & 0.019  \\
					\bottomrule
				\end{tabular}
			\end{sc}
		\end{small}
	\end{center}
	\vskip -0.1in
\end{table}

\begin{table}[t]
	\caption{Action Shapley (AS) for data center cooling management case study for two different RL algorithms, SAC-PID and PPO-PID}
	\label{tab:as_c5}
	\vskip 0.15in
	\begin{center}
		\begin{small}
			\begin{sc}
				\begin{tabular}{lcccr}
					\toprule
					Data set & AS SAC-PID & AS PPO-PID \\
					\midrule
					$c_1$   & 0.9 & 0.88      \\
					$\bf{c_2} $   & \bf{0.93} & \bf{0.94}      \\
					$c_3 $   & 0.91 & 0.92   \\
					
					$c_4$   & 0.63 & 0.62     \\
					$c_5$   & 0.52 & 0.51      \\
					$c_6$   & 0.67 & 0.67  \\
					
					$c_7$   & 0.74 & 0.75 \\
					$c_8$   & 0.75 & 0.73 \\
					$c_9$   & 0.72 & 0.74\\
					
					$c_{10}$   & 0.08 & 0.07 \\
					$c_{11}$   & 0.16 & 0.17 \\
					$c_{12}$   & 0.22 & 0.21 \\
					\bottomrule
				\end{tabular}
			\end{sc}
		\end{small}
	\end{center}
	\vskip -0.1in
\end{table}

\subsection{Action Shapley Values for Training Data}

Table \ref{tab:as_c1} displays the two Action Shapley  values corresponding to two RL algorithms SAC-PID and PPO-PID, and  five distinct training data points, denoted as $\langle a_1, a_2, a_3, a_4, a_5 \rangle$, within the VM right-sizing case study. Notably, it was identified that the training data points $a_3$, $a_4$, and $a_5$ are indispensable. Between two remaining actions, $a_2$ exhibits the highest Action Shapley values, specifically -1.84 and -1.81, respectively for SAC-PID and PPO-PID. The global cut-off cardinality, denoted as $\theta$, is set at 4, resulting in $P_{comp}=50\%$ as defined by Equation \ref{eq:comp_effort}. It is noteworthy that both SAC-PID and PPO-PID demonstrate congruent Action Shapley values and identical cut-off cardinality values.

Table \ref{tab:as_c2} displays the two Action Shapley  values corresponding to two RL algorithms SAC-PID and PPO-PID, and  five distinct training data points, denoted as $\langle w_1, w_2, w_3, w_4, w_5 \rangle$, within  the load balancing case study.  $w_3$ exhibits the highest Action Shapley values, specifically 8.31 and 8.27, respectively for SAC-PID and PPO-PID. The global cut-off cardinality, denoted as $\theta$, is set at 3, resulting in  $P_{comp}=75\%$ as defined by Equation \ref{eq:comp_effort}. It is noteworthy that both SAC-PID and PPO-PID demonstrate congruent Action Shapley values and identical cut-off cardinality values.

Table \ref{tab:as_c3} displays the two Action Shapley values corresponding to two RL algorithms SAC-PID and PPO-PID, and  six distinct training data points, denoted as $\langle p_1, p_2, p_3, p_4, p_5, p_6 \rangle$, within  the database tuning case study.  Notably, it was observed that the training data point $p_1$ is deemed indispensable, whereas the remaining five training data points are considered dispensable.  $p_3$ exhibits the highest Action Shapley values, specifically 3.42 and 3.37, respectively for SAC-PID and PPO-PID. The global cut-off cardinality, denoted as $\theta$, is set at 4, resulting in $P_{comp}=75\%$ as defined by Equation \ref{eq:comp_effort}. It is noteworthy that both SAC-PID and PPO-PID demonstrate congruent Action Shapley values and identical cut-off cardinality values.

Table \ref{tab:as_c4} displays the two Action Shapley  values corresponding to two RL algorithms SAC-PID and PPO-PID, and  15 distinct training data points, denoted as $\langle r_1, r_2, r_3, r_4, r_5, r_6, r_7, r_8, r_9, r_{10}, r_{11}, r_{12}, r_{13}, r_{14}, r_{15} \rangle$, within  the K8s case study.  Notably,  $r_{13}$ exhibits the highest Action Shapley values, specifically 0.77 for both SAC-PID and PPO-PID. The global cut-off cardinality, denoted as $\theta$, is set at 5, resulting in $P_{comp}=99.9\%$ as defined by Equation \ref{eq:comp_effort}. It is noteworthy that both SAC-PID and PPO-PID demonstrate congruent Action Shapley values and identical cut-off cardinality values.

Table \ref{tab:as_c5} displays the two Action Shapley  values corresponding to two RL algorithms SAC-PID and PPO-PID, and  12 distinct training data points, denoted as $\langle c_1, c_2, c_3, c_4, c_5, c_6, c_7, c_8, c_9, c_{10}, c_{11}, c_{12} \rangle$, within  the data center cooling management case study.  Notably,  $c_{2}$ exhibits the highest Action Shapley values, specifically 0.93 for SAC-PID and 0.94 for PPO-PID. The global cut-off cardinality, denoted as $\theta$, is computed to be 6, resulting in $P_{comp}=98.44\%$ as defined by Equation \ref{eq:comp_effort}. It is noteworthy that both SAC-PID and PPO-PID demonstrate congruent Action Shapley values and identical cut-off cardinality values.

\begin{table}[t]
	\caption{\centering{This table summarizes the number of training data points and related parameters for four different case studies.}}
	\label{tab:choice}
	\begin{center}
		\begin{tabular}{lccccc}
			\toprule
			 Case Study & VM & Load  & DB & K8s & Cooling \\
			\midrule
			
			Training Data &  5  &  5 & 6 & 15  & 12  \\
			Cut-off &  4  &  3 & 4 & 5  & 6 \\
			Indispensable &  3  &  0 & 1 & 0 & 2  \\
			DF &  1  &  3 & 3 & 5 & 4   \\
			Choice Set & 2 & 10 & 10 & 3003 & 210 \\
			\bottomrule
			
		\end{tabular}
	\end{center}
\end{table}

\subsection{Validation of Action Shapley Based Training Data Selection Policy}

In Methodology section, we introduced a training data selection policy: the optimal training dataset should include the maximum number of training data points, up to the computed global cut-off cardinality, with the highest average Action Shapley value. The selection process is further refined when indispensable data points are considered. For instance, Table \ref{tab:choice} provides a demonstration in this context. In VM right-sizing case study, we need to choose a single data point from 2 dispensable data points, therefore, we have ${2 \choose 1} = 2$ available options. In both the load balancing and database tuning cases, the task involves selecting 3 data points out of 5 dispensable data points, resulting in ${5 \choose 3} = 10$ potential choices for these specific case studies. In the K8s management case study, where the selection involves 5 data points out of 15 dispensable data points, there are ${15 \choose 5} = 3,003$ possible choices. In data center cooling management case study, we need to choose four data points from 10 possible data points, therefore, we have ${10 \choose 4} = 210$ available options.

\begin{figure}[ht]
	\vskip 0.2in
	\begin{center}
		\centerline{\includegraphics[width=\columnwidth]{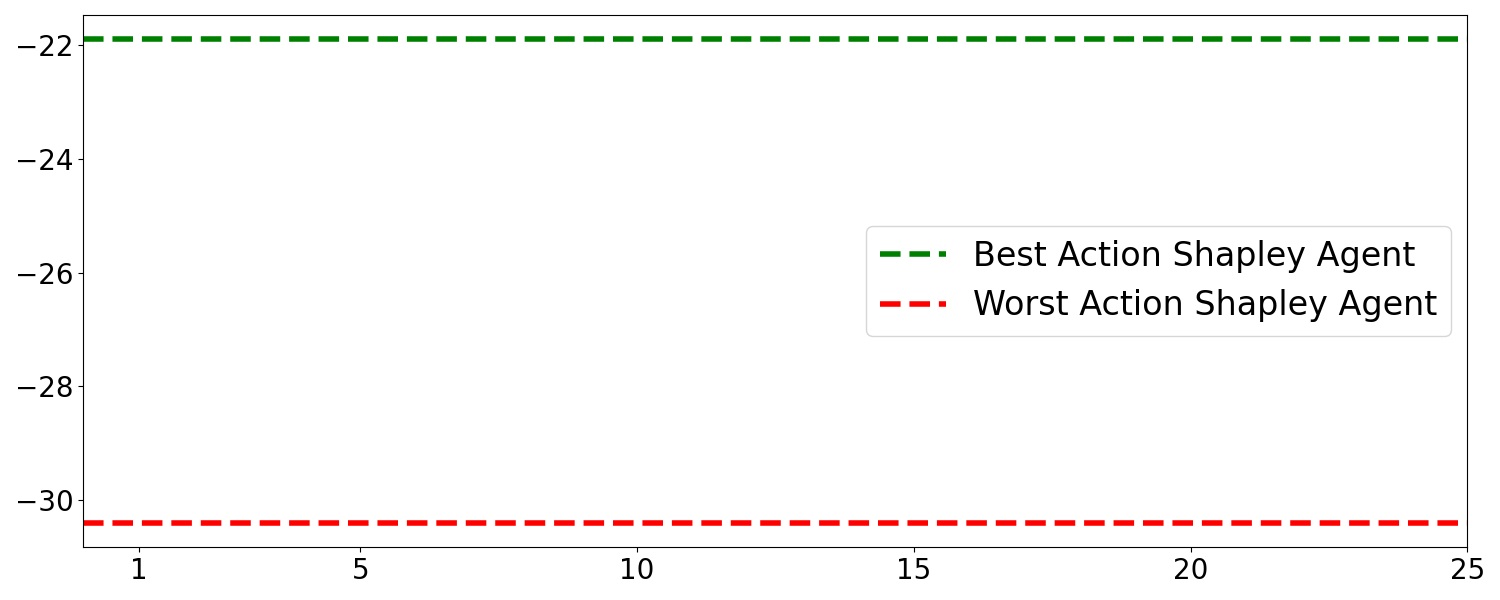}}
			\caption{This plot compares the cumulative reward scores for the best Action Shapley agent vs the worst Action Shapley agent for the VM right-sizing case study. }
		\label{fig:hyp_testing_vm_rightsizing} 
	\end{center}
	\vskip -0.2in
\end{figure}

Our empirical validation process involving five case studies consists of two steps. Firstly, we assess whether the best Action Shapley agent, generated from the training set with the highest average Action Shapley value and an element count matching the cut-off cardinality, achieves a higher cumulative reward compared to the worst Action Shapley agent. The worst Action Shapley agent is derived from the training set with the lowest average Action Shapley value and the same number of elements as the cut-off cardinality. Subsequently, we investigate whether the best Action Shapley agent consistently outperforms the majority of other agents. To answer this question, we conduct a series of 25 episodes, each involving multiple random selections. Each random selection is characterized by a training data set size equivalent to the cut-off cardinality. In light of the comparable Action Shapley values generated by both SAC-PID and SAC-PPO, we choose to utilize SAC-PID-based Action Shapley values in this section for the sake of convenience and without sacrificing generality.

\begin{figure}[ht]
	\vskip 0.2in
	\begin{center}
		\centerline{\includegraphics[width=\columnwidth]{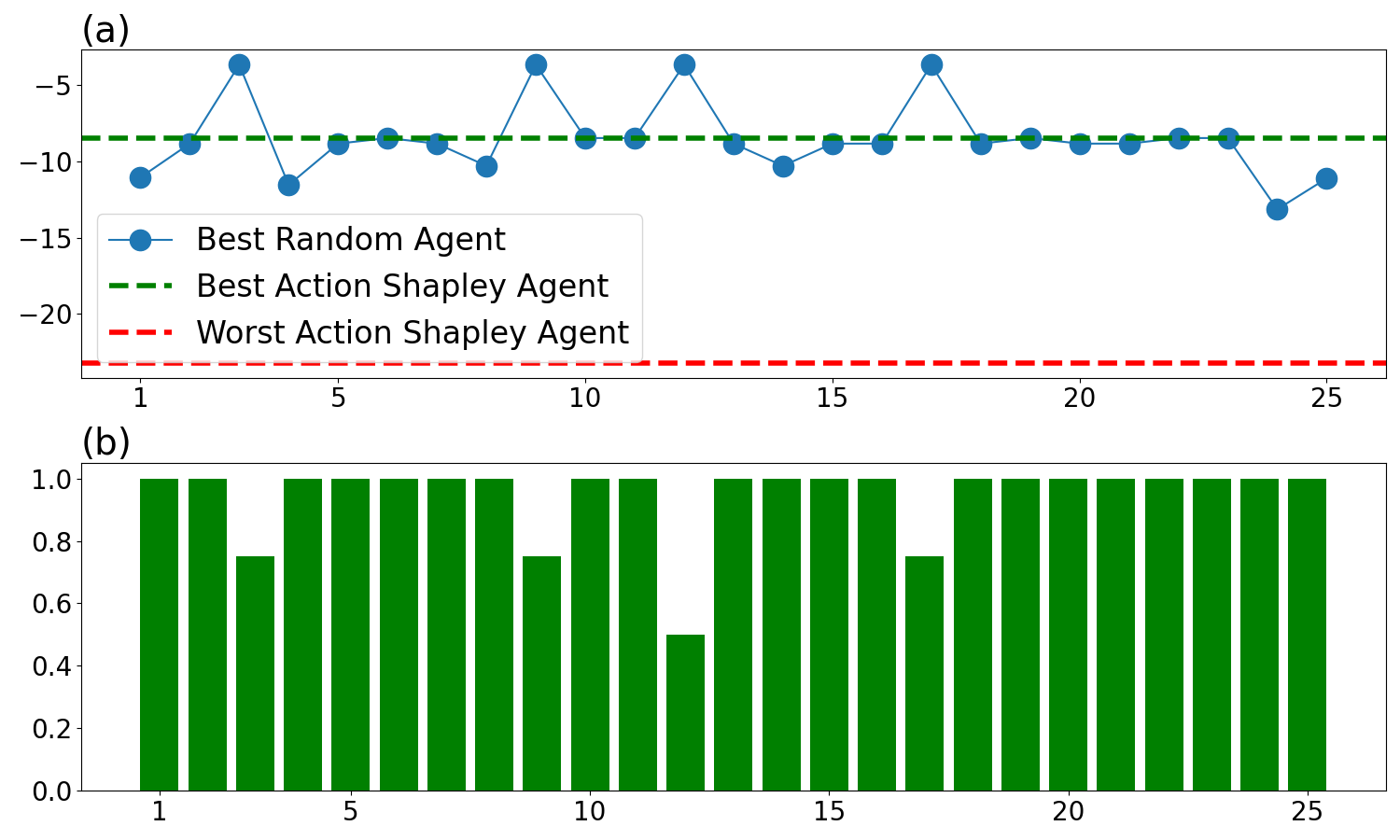}}
			\caption{Validation of Action Shapley based selection policy for the load balancing case study. (a) Comparisons of cumulative rewards among the best Action Shapley agent, the worst Action Shapley agent, and the best of 4 random training action sets for 25 different episodes.  (b) Fractions of agents based on 4 random training datasets with lower cumulative rewards than that of the best Action Shapley agent for 25 episodes.}
		\label{fig:hyp_testing_load_balance} 
	\end{center}
	\vskip -0.2in
\end{figure}

For the VM right-sizing case study, our options for training data sets are limited to two. Consequently, our validation process is distilled into a comparative analysis between the best Action Shapley agent and the worst Action Shapley agent. Illustrated in Figure \ref{fig:hyp_testing_vm_rightsizing}, the cumulative reward score for the superior Action Shapley agent registers at $-21.9$, while that for the inferior counterpart is recorded at $-30.4$. This discernible discrepancy in cumulative rewards substantiates the efficacy of the proposed training action selection policy.

In the context of the load balancing case study, a judicious selection of three training actions from a pool of five is imperative. The cumulative rewards, as depicted in Figure \ref{fig:hyp_testing_load_balance}(a), elucidate a noteworthy performance disparity between the best and worst Action Shapley agents, with respective scores of $-8.49$ and $-23.2$. Notably, the optimal Action Shapley agent exhibits comparable efficacy to the highest achiever among randomly chosen sets across 25 episodes. Figure \ref{fig:hyp_testing_load_balance}(b) further presents the proportion of randomly assembled training action sets resulting in agents with cumulative rewards surpassing that of the premier Action Shapley agent. The discernible trend reveals a mere 5 out of 100 randomly derived agents outperforming the optimal Action Shapley agent. Consequently, a confident assertion can be made that this case study effectively validates the viability of our chosen training data selection policy.

\begin{figure}[ht]
	\vskip 0.2in
	\begin{center}
		\centerline{\includegraphics[width=\columnwidth]{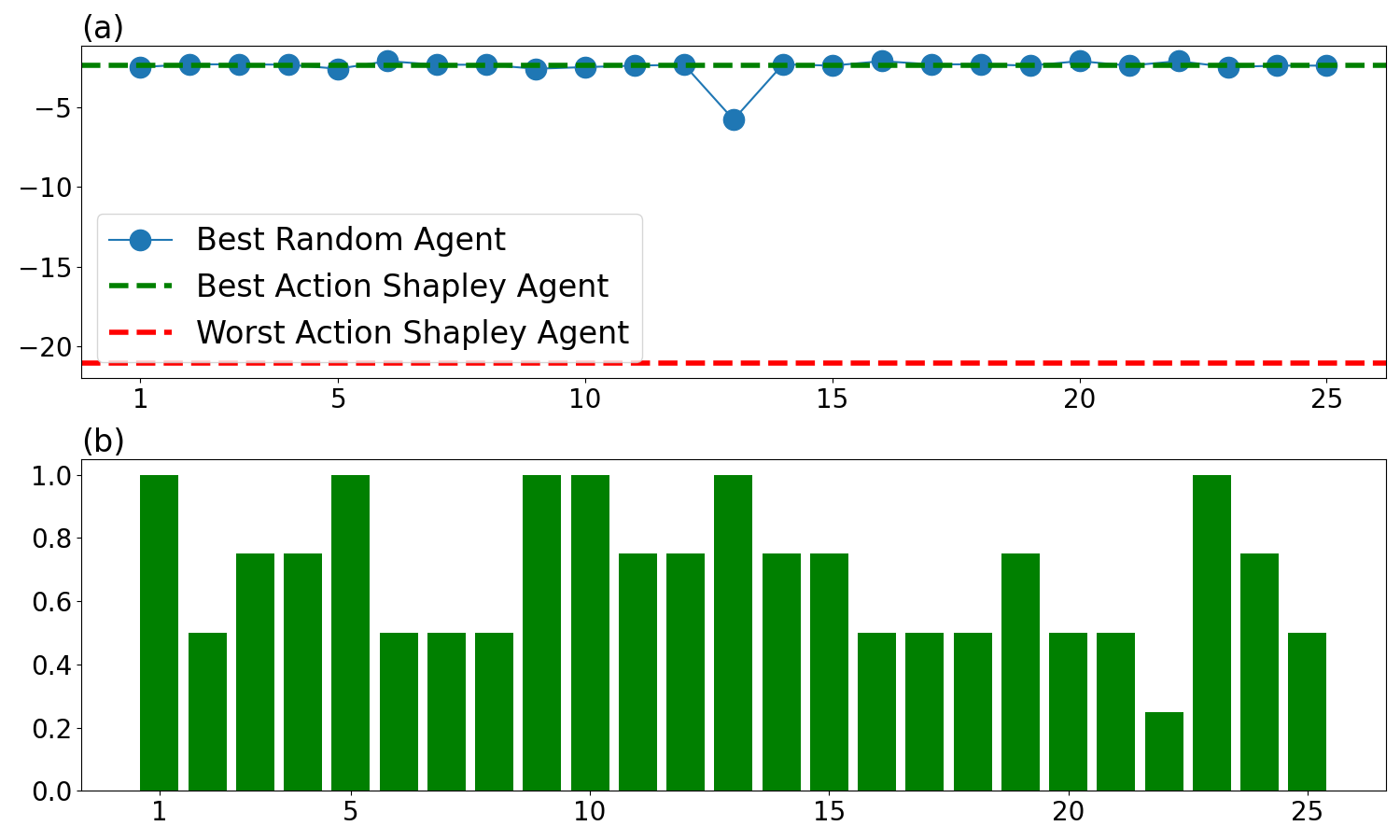}}
		\caption{Validation of Action Shapley based selection policy for database tuning. (a) Comparisons of cumulative rewards among the best Action Shapley agent, the worst Action Shapley agent, and the best of 4 random training action sets for 25 different episodes.  (b) Fractions of agents based on 4 random training datasets with lower cumulative rewards than that of the best Action Shapley agent for 25 episodes.}
		\label{fig:hyp_testing_database} 
	\end{center}
	\vskip -0.2in
\end{figure}

For the database tuning case study, we faced the task of selecting four training datasets out of a pool of six. As depicted in Figure \ref{fig:hyp_testing_database}(a), the cumulative reward for the best Action Shapley agent, standing at $-2.42$, surpasses that of the worst Action Shapley agent, which is $-21$. Additionally, the figure illustrates that the performance of the best Action Shapley agent is comparable to the top-performing agent derived from four randomly selected training action sets across each of the 25 episodes. Figure \ref{fig:hyp_testing_database}(b) presents the fraction of randomly chosen training action sets that yield an agent with a cumulative reward lower than that of the best Action Shapley agent. While 31 out of 100 random selections perform better than the best Action Shapley agent, it's crucial to note that this difference is marginal, as evidenced by the top subplot comparing the best Action Shapley performance with the top performers in each of the 25 episodes. Consequently, we assert that this case study serves as a validation of our training data selection policy.

\begin{figure}[ht]
	\vskip 0.2in
	\begin{center}
		\centerline{\includegraphics[width=\columnwidth]{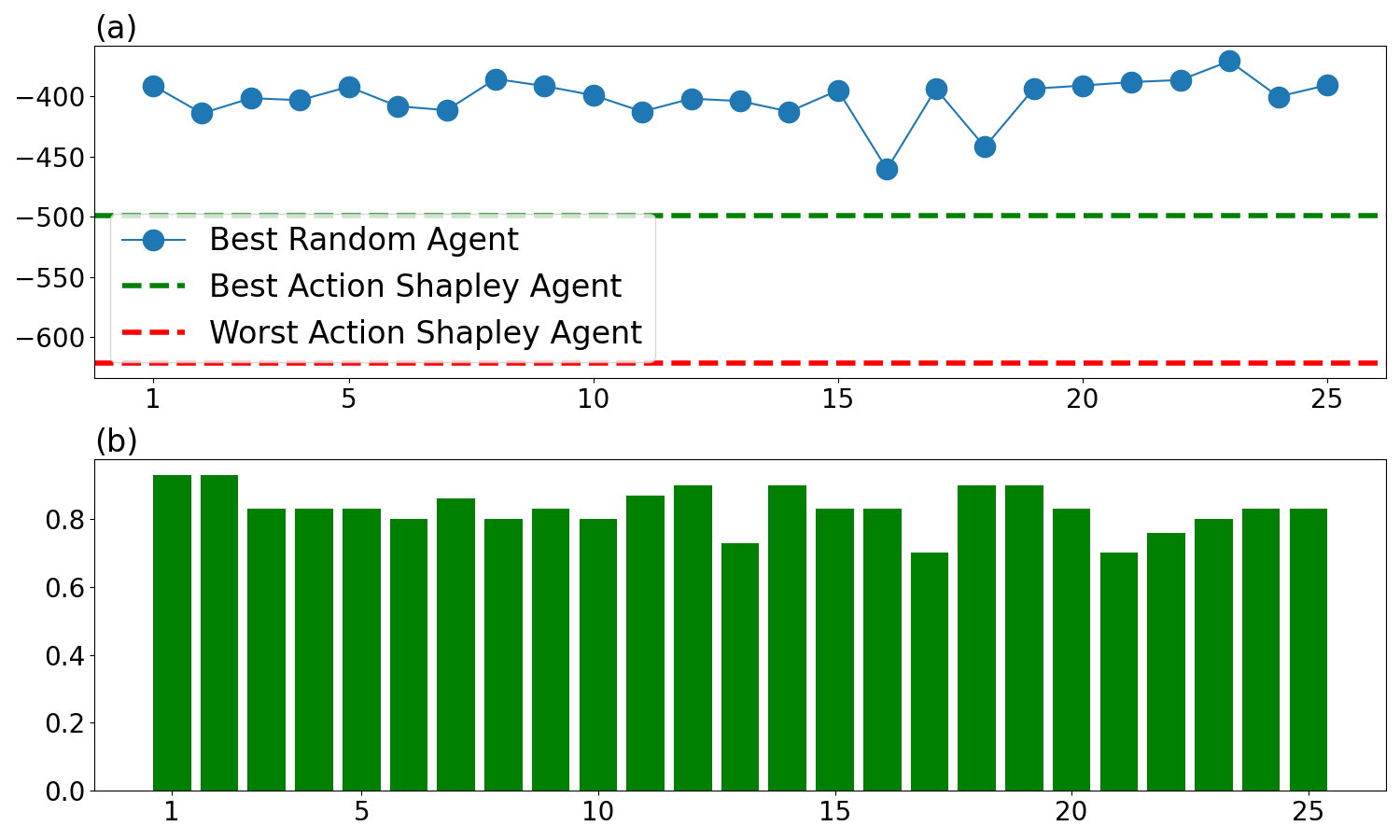}}
		\caption{Validation of Action Shapley based selection policy for K8s management. (a) Comparisons of cumulative rewards among the best Action Shapley agent, the worst Action Shapley agent, and the best of 30 random training action sets for 25 different episodes.  (b) Fractions of agents based on 30 random training datasets with lower cumulative rewards than that of the best Action Shapley agent for 25 episodes.}
		\label{fig:hyp_testing_K8s} 
	\end{center}
	\vskip -0.2in
\end{figure}

In our Kubernetes management case study, we had a pool of 3,003 options for selecting the training data set. Figure \ref{fig:hyp_testing_K8s}(a) displays the cumulative reward, indicating that the best Action Shapley agent achieved a score of $-499$, surpassing the worst Action Shapley agent, which scored $-621$. Furthermore, it demonstrates that the best Action Shapley agent's performance is on par with the top-performing agent from 30 randomly selected sets across each of the 25 episodes. Figure \ref{fig:hyp_testing_K8s}(b) presents the proportions of random training data sets resulting in an agent with a cumulative reward lower than that of the best Action Shapley agent. Notably, in this figure, it's observed that 125 out of 750 random selections outperform the best Action Shapley agent in terms of cumulative reward. However, it is crucial to highlight that despite this, the top performers from these random selections exhibit performance levels comparable to the best Action Shapley agent. Therefore, we confidently assert that this case study provides validation for our training data selection policy.

\begin{figure}[ht]
	\vskip 0.2in
	\begin{center}
		\centerline{\includegraphics[width=\columnwidth]{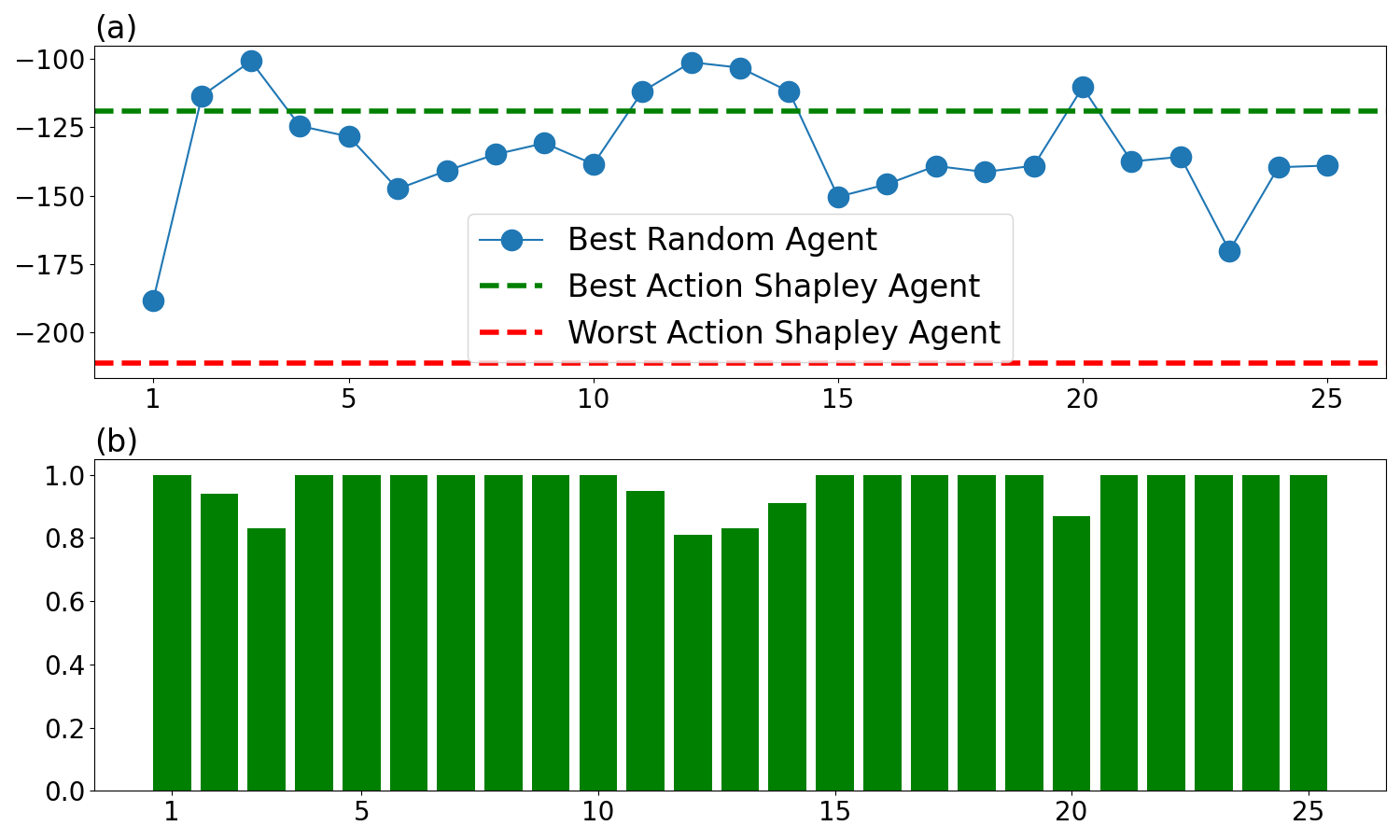}}
		\caption{Validation of Action Shapley based selection policy for data center cooling management. (a) Comparisons of cumulative rewards among the best Action Shapley agent, the worst Action Shapley agent, and the best of 10 random training action sets for 25 different episodes.  (b) Fractions of agents based on 10 random training datasets with lower cumulative rewards than that of the best Action Shapley agent for 25 episodes.}
		\label{fig:hyp_testing_cooling} 
	\end{center}
	\vskip -0.2in
\end{figure}

In the data center cooling management case study, we had a pool of 210 options for selecting the training data set. Figure \ref{fig:hyp_testing_cooling}(a) displays the cumulative reward, indicating that the best Action Shapley agent achieved a score of $-119$, surpassing the worst Action Shapley agent, which scored $-211$. Furthermore, it demonstrates that the best Action Shapley agent's performance is on par with the top-performing agent from 10 randomly selected sets across each of the 25 episodes. Figure \ref{fig:hyp_testing_cooling}(b) presents the proportions of random training data sets resulting in an agent with a cumulative reward lower than that of the best Action Shapley agent. Notably, in this figure, it's observed that 9 out of 250 random selections outperform the best Action Shapley agent in terms of cumulative reward. Therefore, we can confidently assert that this case study provides validation for our training data selection policy.

\subsection{Comparison to Baseline Results}

It is imperative to assess the efficacy of Action Shapley by benchmarking it against a baseline study that assumes the utilization of all available training data points for training the environment model. Specifically, in the VM right-sizing case study, the baseline study incorporates 5 training data points instead of the specified cutoff cardinality of 4. In the load balancing case study, the baseline study utilizes 5 training data points, deviating from the specified cutoff cardinality of 3. Similarly, in the database tuning case study, the baseline study integrates 6 training data points rather than the stipulated cutoff cardinality of 4. In the K8s management case study, the baseline study employs 15 training data points, exceeding the defined cutoff cardinality of 5. Lastly, in the data center cooling management case study, the baseline study employs 12 training data points, exceeding the cutoff cardinality of 6.

The second evaluative dimension considers the cumulative reward attained by both the baseline agent and the optimal Action Shapley agent. Notably, in four of the five case studies (VM right-sizing, load balancing, K8s management, and data center cooling management), the optimal Action Shapley agent demonstrates a significant performance advantage over the baseline agent. Specifically, for VM right-sizing, the values are -37.7 compared to -21.9; for load balancing, -9.1 compared to -8.49; for K8s management, -561.4 compared to -499; and for data center cooling management, -205.1 compared to -119.

\section{Related Work}

 Within the realm of machine learning, the Shapley value have found application in data valuation \citep{data_shapley}, model interpretability \citep{sundararajan2020many}, and feature importance \citep{SHAP}. The Shapley value-based model explanation methodology falls under Additive Feature Attribution Methods \citep{additive}, which includes other methods such as LIME \citep{LIME}, DeepLIFT \citep{deeplift}, and Layer-Wise Relevance Propagation \citep{layer_wise_prop}.

In addressing the intricate challenge of training data selection in reinforcement learning (RL), compounded by the deadly triad \citep{sutton2018reinforcement}, researchers have explored diverse perspectives, including hindsight conditioning \citep{hindsight_conditioning}, return decomposition \citep{return_decomposition}, counterfactual multi-agent policy gradients \citep{counterfactual_ma}, corruption robustness \citep{corruption_robustness}, optimal sample selection \citep{optimal_sampling}, active learning \ \citep{kwik}, minimax PAC \citep{azar_sample_complexity}, $\epsilon$-optimal policy \citep{sidford_sample_complexity}, regret minimization \citep{jin_sample_complexity}, and statistical power \citep{reproducibility}.

The literature extensively explores the use of Shapley values for crediting agents in multi-agent RL \citep{rl_shapley}. Various methods for estimating Shapley value feature attributions are proposed in \citep{nature2023algorithms}. Notably, there is a lack of dedicated studies on applying the Shapley value to select training data for reinforcement learning environment model training.

\section{Conclusion}

This paper validates Action Shapley as a training data selection metric for reinforcement learning with five empirical case studies. To address the inherent exponential time complexity, a randomized algorithm is proposed for computing Action Shapley.  The motivation behind this research stems from the critical role that life cycle management of  training data plays in distinguishing the performance of RL agents. We anticipate that the adoption of Action Shapley will facilitate real-world applications of model-based reinforcement learning in dynamic control systems.

\bibliography{iclr2026_conference}
\bibliographystyle{iclr2026_conference}

\newpage
\appendix
\section{Appendix}
\begin{algorithm}
	\caption{Algorithm for Action Shapley Computation for a Training Action, $k$}
	\label{alg:ASC}
	\textbf{Input}: the total number of training data points: $n$; the set of all training data points: $\mathcal{D}$; RL algorithm used: $\mathcal{A}$; and valuation function for a training action subset without $k$: $\mathcal{U}(\mathcal{D}\backslash \{k\}; \mathcal{A}).$\\
	\textbf{Output}: Action Shapley value: $\phi_{k}$; cut-off cardinality: $\theta_k$ for data point $k$.\\
	
	\textbf{Parameter}: arbitrary constant: $C_f$;  error bound: $\epsilon$.\\
	\textbf{Variable}: training subset cardinality index: ${i}$; accumulator for marginal contributions for different subsets: ${sum}$; training subset: ${d}$; failure memoization: ${mem}$; termination indicator: $flag$\\
	
	\begin{algorithmic}[1] %[1] enables line numbers
		\STATE let $flag=0$ 
		\STATE let $i=n-1$
		\STATE let $sum=0$ 
		\STATE let $\theta_k=1$ 
		
		\WHILE{iterate all cardinality greater than 1: $ i > 1$}
		\STATE let $mem=0$
		\WHILE{iterate all the sets of cardinality $i$ : $ d \in \mathcal{D}_i \backslash \{k\} $}
		\IF {($\mathcal{U}(d\cup\{k\})$ is $null$) $\lor$ ($\mathcal{U}(d)$ is $null$)}
		\STATE $sum=sum$
		\STATE $mem=mem+1$
		\IF {$mem ==  \epsilon$}
		\STATE $\theta_k = i + 1$
		\STATE $flag=1$
		\STATE \textbf{break} \COMMENT{get out of the inner loop (line 7)}
		\ENDIF
		\ELSE
		\STATE  $sum = sum + C_f \frac{\mathcal{U}(d \cup \{k\}) -  \mathcal{U}(d)}{\binom{n-1}{\vert{d}\vert}}$
		\ENDIF
		\ENDWHILE
		\IF {$flag==1$}
		\STATE \textbf{break} \COMMENT{get out of the outer loop (line 5)}
		\ELSE
		\STATE $i=i-1$
		\ENDIF
		
		\ENDWHILE
		\STATE $\phi_{k}=sum$
		\STATE \textbf{return} $\phi_{k}, \theta_k$
	\end{algorithmic}
\end{algorithm}

\end{document}